\documentclass[10pt,twocolumn,letterpaper]{article}

\usepackage[pagenumbers]{iccv} 

\usepackage{amsmath}
\usepackage{mathtools} 
\usepackage[linesnumbered,ruled,vlined,algo2e]{algorithm2e}
\usepackage{algorithm}
\usepackage{algorithmic}
\usepackage{multirow}
\usepackage{soul}
\usepackage{wrapfig}

\newcommand{\boldstart}[1]{\vspace{3pt}\noindent\textbf{#1}}
\newcommand{\Boldstart}[1]{\vspace{6pt}\noindent\textbf{#1}}

\definecolor{iccvblue}{rgb}{0.21,0.49,0.74}
\usepackage[pagebackref,breaklinks,colorlinks,allcolors=iccvblue]{hyperref}

\def\paperID{***} 
\def\confName{ICCV}
\def\confYear{2025}

\title{
Active View Selector: Fast and Accurate Active View Selection with Cross Reference Image Quality Assessment
}

\author{
Zirui Wang$^\dagger$ \and
Yash Bhalgat$^\star$ \and
Ruining Li$^\star$ \and
Victor Adrian Prisacariu$^\dagger$ \and
$^\dagger$Active Vision Lab \quad $^\star$Visual Geometry Group \\
University of Oxford \\
{\tt\small \{ryan, yashsb, ruining, victor\}@robots.ox.ac.uk}
}

\begin{document}
\maketitle

\begin{abstract}
We tackle active view selection in novel view synthesis and 3D reconstruction. Existing methods like FisheRF and ActiveNeRF select the next best view by minimizing uncertainty or maximizing information gain in 3D, but they require specialized designs for different 3D representations and involve complex modelling in 3D space.
Instead, we reframe this as a 2D image quality assessment (IQA) task, selecting views where current renderings have the lowest quality. Since ground-truth images for candidate views are unavailable, full-reference metrics like PSNR and SSIM are inapplicable, while no-reference metrics, such as MUSIQ and MANIQA, lack the essential multi-view context.
Inspired by a recent cross-referencing quality framework CrossScore, we train a model to predict SSIM within a multi-view setup and use it to guide view selection.
Our cross-reference IQA framework achieves substantial quantitative and qualitative improvements across standard benchmarks, while being agnostic to 3D representations, and runs 14-33 times faster than previous methods. Project page: \href{https://avs.active.vision/}{\texttt{https://avs.active.vision/}}
\end{abstract}
    
\begin{figure}[t]
    \centering
    \includegraphics[width=1\columnwidth]{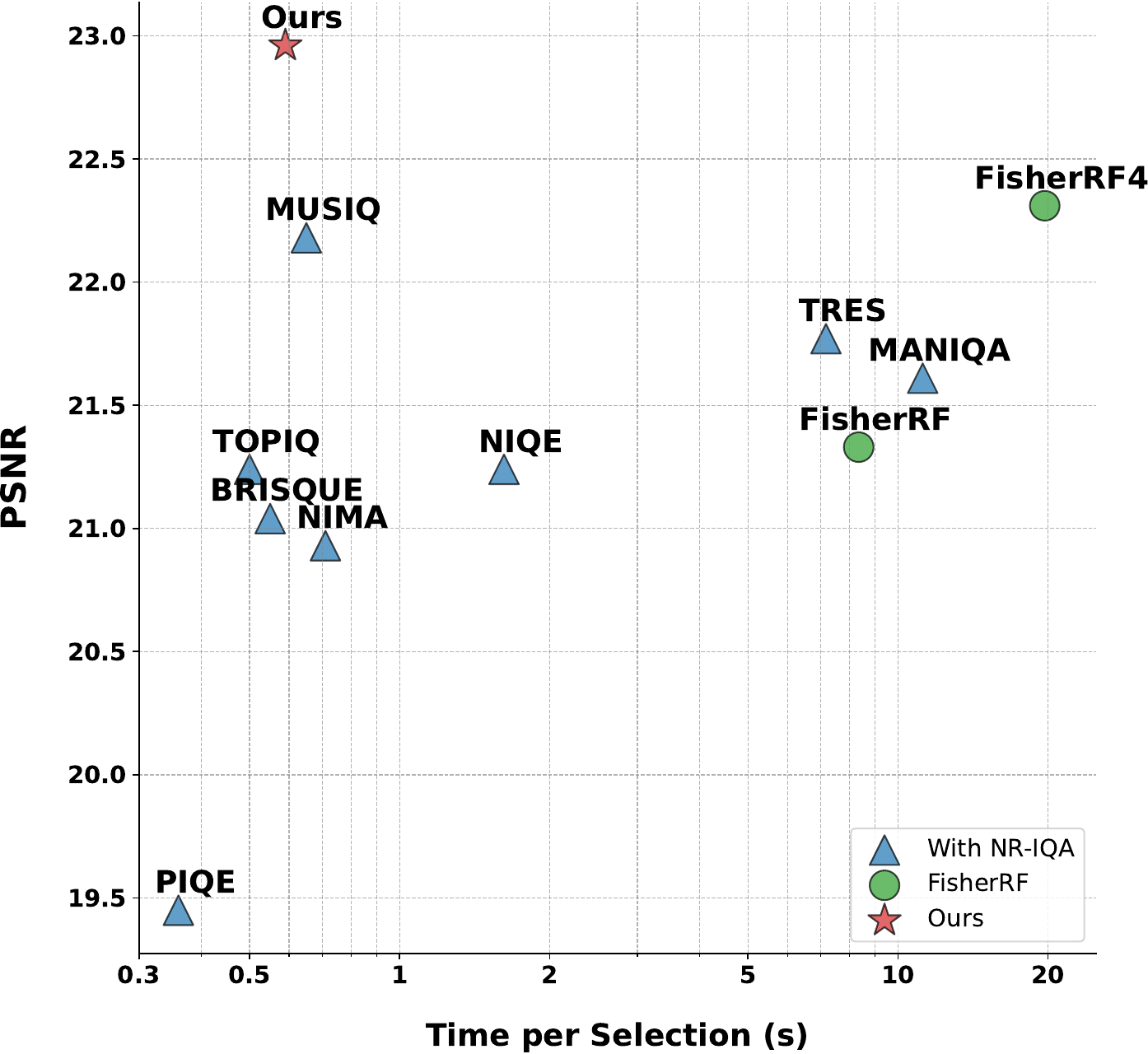}
    \caption{
    \textbf{
    View selection time in seconds ($\downarrow$) \vs NVS quality measured by PSNR ($\uparrow$) on the \emph{Garden} scene from Mip-NeRF360 dataset.~\cite{barron2021mip}.}
    Our method achieves a $14\times$ speedup over the state-of-the-art view selection method, FisherRF~\cite{jiang2025fisherrf}, and a $33\times$ speedup over its batched variant, FisherRF4, while achieving improved NVS quality. Notably, several no-reference IQA-based approaches also emerge as strong baselines for this task.
    }
    \vspace{-0.3cm}
    \label{fig:teaser}
\end{figure}

\section{Introduction}
\label{sec:intro}

3D representations such as NeRF~\cite{mildenhall2020nerf} and Gaussian Splatting~\cite{kerbl2023gs} can achieve remarkable photorealism in 3D scene reconstruction. A flurry of research has reduced the latency of radiance field rendering by several orders of magnitude~\cite{barron2021mip, muller2022instant, chen2022tensorf, kerbl2023gs}.
However, these 3D reconstruction methods require a large number of posed views to achieve high accuracy.
While a substantial body of work has focused on sparse view reconstruction~\cite{yu2021pixelnerf, hong2023lrm, tang2024lgm, li2024instant3d, chan2023generative, wu2023magicpony, li2024learning, jakab2024farm3d, szymanowicz2024splatter} they face challenges in incorporating additional views to refine the reconstruction further.

In contrast to \textit{passively} performing 3D reconstruction with a fixed set of images, another approach is \textit{active} reconstruction, where an active view selection system suggests the next optimal viewpoints to maximize reconstruction quality.
This is particularly important when image acquisition is time- or power-consuming, such as in path planning and space exploration in robotic applications.

To determine where to select the next view, previous works consider uncertainty~\cite{pan2022activenerf} or information gain~\cite{jiang2025fisherrf} in 3D. 
Two key shortcomings of these approaches are \textit{runtime efficiency} and \textit{architecture dependence}.
For example, FisherRF~\cite{jiang2025fisherrf} calculates the information gain using Fisher Information, which requires constructing a Hessian matrix with dimensions matching the scene parameters, typically over $200$ million for Gaussian Splats.
This selection process remains prohibitively slow (see~\cref{fig:teaser}).
Moreover, since the information gain or uncertainty is determined in 3D space, it inevitably relies on 3D representations 
and network structures, leading to challenges when adapting these methods to apply to various 3D representations, for example, NeRF, 3DGS, SDF-based, or voxel-based representations.

In this study, we challenge the need for 3D uncertainty and 3D information gain based view selection algorithms.
We observe that existing approaches, regardless of how they quantify informativeness, tend to select the next view from a viewpoint where the rendering of an initial reconstruction is perceptually lower in quality.
This insight drives us to adopt the visual quality of a 2D rendering as a more efficient proxy metric for guiding view selection.

Our experiments with no-reference (NR) IQA metrics, which are designed to evaluate image quality without ground-truth images, reveal three benefits of IQA-based view selection strategy: \textbf{\textit{effectiveness, lightweight implementation, and representation agnosticism}}.
Specifically, we found several IQA metrics match or even outperform the effectiveness of FisherRF on the NVS task, while achieving significantly faster runtimes (see \cref{fig:teaser}).
Moreover, since IQA metrics require images only, they are agnostic to 3D representations and network architectures, unlike ActiveNeRF and FisherRF, both of which require considerable modifications when migrating between NeRF and 3DGS.

To enhance IQA performance, we employ a cross-reference IQA scheme~\cite{wang2024crossscore} to assess image quality in a multi-view setting. This image score guides viewpoint selection by identifying the lowest-quality perspective.
We train the model in a self-supervised manner by constructing a dataset of distorted images and SSIM maps, which supervise the IQA predictions. The dataset is generated by fitting radiance fields to scenes and computing SSIM maps between rendered and ground-truth images.

To summarize, we make the following \textbf{contributions}:
\textbf{First}, we identify the effectiveness and simplicity when using 2D image quality assessment for view selection. This IQA based view selection strategy is agnostic to 3D representations and network structure, significantly reduced complexity in view selection systems.
\textbf{Second}, within 2D IQA scheme, we further designed a low latency cross-reference IQA model that incorporates multi-view priors, enhancing view selection outcomes.
\textbf{Third}, We evaluate our active view selection approach on two key tasks across multiple standard benchmarks, demonstrating that it achieves significantly improved quantitative and qualitative results while reducing latency by $14$-$33\times$.
    
\section{Related Work}
\label{sec:related_work}

\boldstart{Active Learning.}
Active learning attempts to maximize the performance gain of a model while annotating the fewest samples possible~\cite{settles2009activelearningsurvey, ren2021activelearningsurvey}.
Typically, queries are drawn in a greedy fashion according to an informativeness measure.
Researchers have proposed frameworks to evaluate the informativeness of unlabeled instances, such as uncertainty sampling~\cite{lewis1995sequential}, where the instances are prioritized if the model is less certain about how to label them;
query-by-committee~\cite{seung1992query}, where a committee of partially trained models selects the most informative query as the one about which they most disagree;
expected model change~\cite{settles2007multiple}, which queries the instance that would impart the greatest change to the current model if we knew its label;
and estimated error reduction~\cite{roy2001toward}, which selects the instance that brings minimal expected future error.
The active view selection problem, the focus of this work, is an application of active learning.
Our framework, based on 2D IQA, can be broadly viewed as an application of the estimated error reduction approach from the active learning literature.

\boldstart{Radiance Fields.}
Radiance fields have emerged as a popular 3D representation thanks to their flexibility and rendering efficiency.
NeRF~\cite{mildenhall2020nerf} proposes to represent a 3D scene as an MLP that maps a 3D location to its RGB color and occupancy.
Follow-ups replace the MLP with architectures with higher inductive bias, such as a multi-resolution hash table~\cite{muller2022instant} and feature grids~\cite{chen2022tensorf}, making training and rendering of radiance fields more efficient.
Gaussian Splatting~\cite{kerbl2023gs} proposes to represent a scene with Gaussian primitives, which enables real-time rendering.
One key limitation of these radiance field representations is the requirement of many posed views, usually on the order of hundreds.
In this work, we attempt to lift this requirement by introducing a method that actively selects the next best view to maximize reconstruction quality.

\begin{figure*}[t]
    \centering
    \includegraphics[width=1\textwidth]{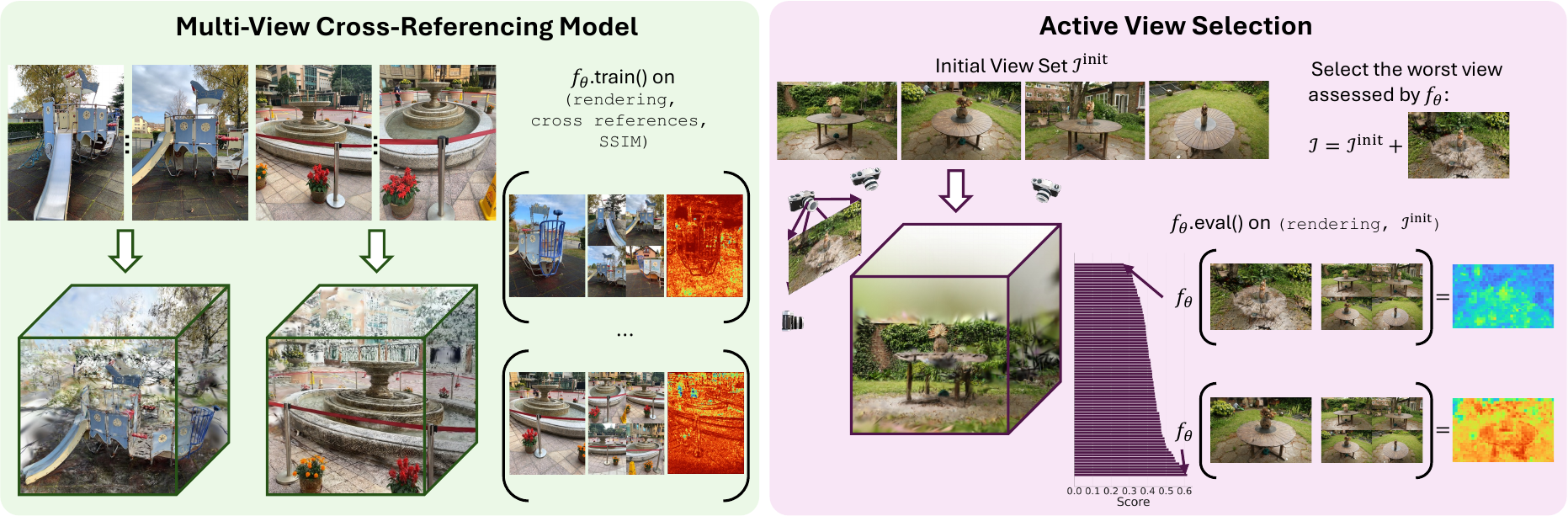}
    \caption{
        \textbf{Method Overview}.
        Our method consists of two main components. First, a lightweight cross-referencing (CR) image quality assessment (IQA) model (\textbf{left}) evaluates a rendered image by comparing it to multiple real images from different viewpoints of the same scene, generating a per-pixel quality map. This model is designed for multi-view novel view synthesis (NVS), where conventional metrics like PSNR and SSIM are inapplicable due to the lack of ground truth images for the novel view. After training on outputs from NVS methods (e.g., Gaussian Splatting, Nerfacto, and TensoRF) across various scenes, it can be applied directly to new real-world scenes in a feed-forward manner.
        The second component is a Gaussian Splatting (GS)-based active view selection system (\textbf{right}). Starting from four views, this system iteratively selects the next best view by: (a) training a GS model on the current view set, (b) rendering candidate viewpoints, (c) evaluating these with our CR-IQA model, (d) selecting the view with the lowest quality, and (e) repeating the process.
    }
    \label{fig:method}
    \vspace{-0.2cm}
\end{figure*}

\boldstart{Active View Selection.}
Most related to ours is the work that combine radiance field optimization with active learning.
ActiveNeRF~\cite{pan2022activenerf} proposes to task the NeRF MLP to output an uncertainty value in addition to color and occupancy for each 3D location, and to select the view that, when introduced, minimizes posterior variance.
FisherRF~\cite{jiang2025fisherrf} selects the next best view based on information gain computed by Fisher Information.
Other works~\cite{jin2023neu, dhami2023pred, ran2023neurar} extend active view selection to active mapping algorithms, improving the mapping efficiency in embodied robotic systems. They mostly perform trajectory planning based on uncertainty estimates.
In this work, we make the observation that uncertainty and information gain are inefficient for active view selection, as
they require explicit modelling in 3D.
They are also scene-specific and do \emph{not} leverage \emph{experience} in reconstructing diverse scenes.
We show that a multi-view cross-referencing model, pre-trained on hundreds of reconstruction episodes for image quality assessment, can be used as an effective metric for active view selection.

\boldstart{Image Quality Assessment (IQA).}
The computer vision community has designed various IQA metrics to benchmark image quality.
Full-reference metrics including pixel-wise mean squared error (MSE), peak signal-to-noise ratio (PSNR), and structural similarity index measure (SSIM~\cite{wang2004image}) require access to ground-truth images, which are not available at scale, or at all, for certain domains.
To avoid relying on ground-truth data, no-reference evaluation metrics are introduced.
Classical approaches like DIIVINE~\cite{diivine}, BRISQUE~\cite{brisque}, NIQE~\cite{niqe}, and PIQUE~\cite{pique} are designed with handcrafted features and statistical models to capture image distortions and estimate quality.
More recently, CrossScore~\cite{wang2024crossscore} proposes a cross-reference framework, which utilizes ground-truth images from \emph{different} viewpoints as context to better perform IQA.
In this work, we introduce a new framework for active view selection that uses no-reference IQA as a black box, and draws inspiration from CrossScore to learn a cross-referencing prior that further improves the view selection results.    
\section{Method}
\label{sec:method}
In the active view selection problem, we are given an initial set of posed views
$
\mathcal{I}^\text{init} \coloneqq
\left\{
(I_i, p_i)
\right\}_{i=1}^{N}
$
and a set $\mathcal{P}$ of candidate poses $\left\{\hat{p}_j\right\}_{j=1}^M$ from which we can obtain additional views to supervise the scene reconstruction. 
Given a limited budget $m < M$, the goal is to select $m$ views that can most effectively boost the reconstruction quality.
Mathematically, we are looking for an $m$-set $\mathcal{Q} \subseteq \mathcal{P}$ such that
$$
R\Big(\mathcal{I}^\text{init}\cup \left\{(\hat{p}_j, \hat{I}_j)\right\}_{j\in \mathcal{Q}}\Big)
\geq
R\Big(\mathcal{I}^\text{init}\cup \left\{(\hat{p}_j, \hat{I}_j)\right\}_{j\in \mathcal{Q^\prime}}\Big)
$$
for any $\mathcal{Q}^\prime \subseteq \mathcal{P}$ of cardinality $\left|\mathcal{Q}^\prime\right|=m$, where $R$ is the reconstruction process and $\geq$ compares the reconstruction quality (larger is better), which in practice is often measured by image-space metrics such as SSIM~\cite{wang2004image} and PSNR from novel viewpoints.
In other words, we aim to find an optimal set of poses $\mathcal{Q}$ that achieves higher reconstruction quality than any alternative set of poses.
Note that we aim to select the $m$ views \emph{adaptively} during reconstruction, instead of choosing them all at once beforehand.

Previous works that consider this problem build up the $m$ additional views by iteratively selecting the next view that maximizes uncertainty reduction~\cite{pan2022activenerf} or information gain~\cite{jiang2025fisherrf} in 3D space.
These ad-hoc solutions are computationally intensive.
In practice, we have observed that the selected viewpoint often corresponds to the current reconstruction's low-quality regions.

Inspired by this, we propose an alternative approach: at each step, we select the additional view from which the rendering of the current reconstructed scene is the \emph{worst} in quality.
We dub this strategy ``boost where it struggles''.
However, since at view selection time the ground-truth rendering is \emph{not} available, we need to perform assessment of the rendering quality using the views at hand, sometimes called \emph{cross-reference} evaluation~\cite{wang2024crossscore}.
In~\cref{sec:method_crossscore}, we introduce the cross-reference framework we leverage for active view selection.
We then propose a simple and effective way to integrate multi-view cross-referencing image quality assessment into active view selection for Gaussian Splatting~\cite{kerbl2023gs} in~\cref{sec:method_pipeline}.
An overview is provided in~\cref{fig:method}.

\subsection{Multi-View Cross-Reference Image Evaluation}
\label{sec:method_crossscore}

Our cross-reference evaluation framework builds on CrossScore~\cite{wang2024crossscore}, which learns to assess novel view renderings within a multi-view context, \emph{without} requiring ground-truth images for the query view.
It does so by training a neural network $f_\theta$, which receives as input a query image $\hat{I}\in \mathbb{R}^{3\times H\times W}$, together with a number of reference images $\left\{I_k\right\}_{k=1}^K$. These reference images are real captured images of the scene from existing training viewpoints, \emph{excluding} the ground-truth $I$ of $\hat{I}$.
The network $f_\theta$ then outputs an evaluation of $\hat{I}$ in the space of common full-reference metrics such as SSIM$(I, \hat{I})$.
Note that $I$ \emph{is} available when training $f_\theta$ but is \emph{not} available at view selection time.

Following~\cite{wang2024crossscore}, we collect training data for $f_\theta$ in a self-supervised manner.
Specifically, we optimize radiance field representations including 3DGS~\cite{kerbl2023gs}, NeRF~\cite{mildenhall2020nerf}, and its variant~\cite{chen2022tensorf} on a number of scene captures.
For each scene with posed captures $\{(I_i, p_i)\}_{i=1}^N$, during optimization we periodically evaluate radiance field renderings $\{\hat{I}_j\}_{j\in\mathcal{J}}$ from camera viewpoints $\{p_j\}_{j\in \mathcal{J}\subseteq \{1, 2, \cdots, N\}}$ against ground-truth images $\{I_j\}_{j\in\mathcal{J}}$.
We then use the evaluation results to construct training triplets
$$
\big( \hat{I}_j, \quad \left\{I_i\right\}_{i \neq j}, \quad \text{SSIM}(\hat{I}_j, I_j) \big)
$$ 
for the network $f_\theta$,
where $\hat{I}_j$ and the $K$ reference images $\left\{I_i\right\}_{i \neq j}$ constitute the input of $f_\theta$, and the SSIM metric is the target output. The network parameters $\theta$ can then be optimized on these triplets using stochastic gradient descent.

By training on the collected large-scale dataset, $f_\theta$ is exposed to various types of artefacts at different levels when reconstructing diverse scenes with multiple radiance field representations, and it generalizes beyond the scenes used in self-supervised data collection.
At inference time, the network is capable of examining a rendering of a novel scene reconstruction against the available real images from different viewpoints and providing an accurate estimate of its SSIM map without requiring ground truth images.

\subsection{Active View Selection with CR-IQA Models}
\label{sec:method_pipeline}

We propose to use the output of the network $f_\theta$ directly as an inverse informativeness metric for each view in active view selection.
On a high level, the cross-reference image quality assessment (CR-IQA) model $f_\theta$ predicts the estimated error for a given rendering. Assuming that each additional view locally reduces reconstruction error around its viewpoint, selecting the view with the highest error prediction (\ie, lowest SSIM score) aligns with the estimated error reduction framework in the active learning literature~\cite{roy2001toward}.
Our view selection process is further detailed in \cref{algo:view_selection}.

Our selection process incurs only a single (batched) forward pass of a lightweight network taking images as input, ensuring that runtime remains unaffected by increasing scene complexity, in contrast to 3D-based FisherRF~\cite{jiang2025fisherrf}.

\boldstart{Directly Applying CrossScore.}
A straightforward choice for the cross-reference evaluation network $f_\theta$ is the original CrossScore~\cite{wang2024crossscore}.
As shown in \cref{sec:exp}, this approach outperforms the previous state-of-the-art method, FisherRF~\cite{jiang2025fisherrf}, in both novel view synthesis performance and view selection time.

\boldstart{Low-latency Modification.}
The original CrossScore was designed for image quality assessment without regard for runtime constraints in real-time systems.
To enable its use on resource-limited embodied robotic platforms, we developed a low-latency variant. Specifically, we replaced the heavy DINOv2~\cite{oquab2023dinov2} backbone in CrossScore with a lightweight CNN-based backbone, RepViT~\cite{wang2024repvit}, pretrained on ImageNet.
On top of the RepViT-backed image encoder, we apply a transformer-based cross-reference module and a shallow MLP to predict the SSIM map. These components closely follow the design of the original CrossScore.

\boldstart{Model Names.}
We refer to the two variants of our active view selection model as Ours-DINOv2 and Ours-RepViT.
By default, we use Ours-RepViT, as replacing the DINOv2 with RepViT results in only a minor performance drop while doubling view selection speed.
Throughout the paper, references to \textit{Ours} indicate the \textit{Ours-RepViT} variant.

\begin{algorithm}[t]
\caption{Active View Selection}
\label{algo:view_selection}
\KwIn{cross-referencing model $f_\theta$, initial view set $\mathcal{I}^\text{init}$, pool of candidate poses $\mathcal{P}$, and view budget $m$.}
\KwOut{selected $m$ views $\mathcal{Q}$, and final scene reconstruction $g_w$.}
\begin{algorithmic}[1]
    \STATE Initialize $g_w$ as random Gaussians, $\mathcal{Q} = \emptyset$.
    \WHILE{$\left|\mathcal{Q}\right| < m$}{
        \STATE Optimize the reconstruction $g_w$ on the posed view set $\mathcal{I}^\text{cur}\coloneqq \mathcal{I}^\text{init}\cup \left\{(\hat{p}_j, \hat{I}_j)\right\}_{j\in \mathcal{Q}}$;
        \STATE Add the view that has the lowest SSIM to $\mathcal{Q}$: $
        \mathcal{Q} \leftarrow \mathcal{Q} \cup \left\{\arg\min_{ p\in \mathcal{P} \setminus \mathcal{Q}} f_\theta\left(\text{render}\left(g_w, p\right), \mathcal{I}^\text{cur} \right) \right\}
        $
    }
    \ENDWHILE
    \RETURN $\mathcal{Q}$ and $g_w$.
\end{algorithmic}
\end{algorithm}

\section{Experiments}
\label{sec:exp}
We first describe the experimental setup in \cref{sec:exp:setup}, then evaluate view selection strategies on two tasks. For Active Novel View Synthesis (\cref{sec:exp:task_a_active_nvs}), we assess image quality using 3D-based methods like FisherRF~\cite{jiang2025fisherrf} and ActiveNeRF~\cite{pan2022activenerf}, along with IQA metrics such as MANIQA and MUSIQ. For 3D mapping (\cref{sec:exp:task_b_3d_recon}), we benchmark reconstruction accuracy using MASt3R~\cite{leroy2024grounding} and SplaTAM~\cite{keetha2024splatam}. Finally, we analyze runtime across all baselines in \cref{sec:exp:run_time}.

\subsection{Experimental Setup}
\label{sec:exp:setup}

\boldstart{Dataset.}
We use three datasets in our experiments for their photorealism and broad adoption. The first is Map-Free Relocalisation (\textbf{MFR})\cite{arnold2022mapfree}, with 460 outdoor videos at 540×960, adapted for training our image scoring network. The second is \textbf{Mip-NeRF360}\cite{barron2021mip}, offering 9 high-res $360^\circ$ scenes, downscaled to 1066×1600. The third is a 10-video subset of RealEstate10K (\textbf{RE10K})~\cite{zhou2018stereo}, downscaled to 960×540. We train on 348 MFR videos and evaluate on 11 held-out videos, plus test on Mip-NeRF360 and RE10K. More details can be found in the supplementary.

\boldstart{Baselines.}
We group active view selection baselines into two categories. The first uses \emph{3D scene uncertainty}, including ActiveNeRF (re-implemented on 3DGS)~\cite{pan2022activenerf} and FisherRF~\cite{jiang2025fisherrf}, which also includes a batched variant, FisherRF4, selecting 4 views each time. The second group relies on \emph{2D image quality}, using no-reference IQA metrics such as BRISQUE~\cite{brisque}, NIQE~\cite{niqe}, PIQE~\cite{pique}, and newer models like TOPIQ~\cite{chen2024topiq}, TRES~\cite{golestaneh2022tres}, MANIQA~\cite{yang2022maniqa}, MUSIQ~\cite{ke2021musiq}, and NIMA~\cite{talebi2018nima}. We use FisherRF official code and the ~\href{https://github.com/chaofengc/IQA-PyTorch}{PyIQA} library for 2D baselines.

\boldstart{Implementation Detail.}
\textit{Architecture:} 
For Ours-DINOv2, we directly use the pretrained CrossScore model.
For Ours-RepViT, we adopt the lightweight CNN-style RepViT~\cite{wang2024repvit} backbone, using the \textit{m0\_9} variant with 26 layers optimized for speed and efficiency. 
Features from layers 2, 6, 22, and 25 are projected to 256 channels, resized to $(h/14, w/14)$, and summed into a final feature map of size $(256, h/14, w/14)$. Cross-referencing uses a two-layer Transformer Decoder with 256 hidden units. A two-layer MLP with sigmoid outputs pixel-wise scores in $[0,1]$.
\textit{Training:} Following CrossScore~\cite{wang2024crossscore}, we train Ours-RepViT by cropping input images to 518×518 and sampling five reference views per query. We train for 20,000 steps using AdamW~\cite{loshchilov2018decoupled} at a constant 5e-4 learning rate on two 24GB GPUs, with batch size 64 per GPU. Total training takes about 13 hours.

\subsection{Task A: Active Novel View Synthesis}
\label{sec:exp:task_a_active_nvs}

\begin{figure*}[t]
    \centering
    \includegraphics[width=1\textwidth]{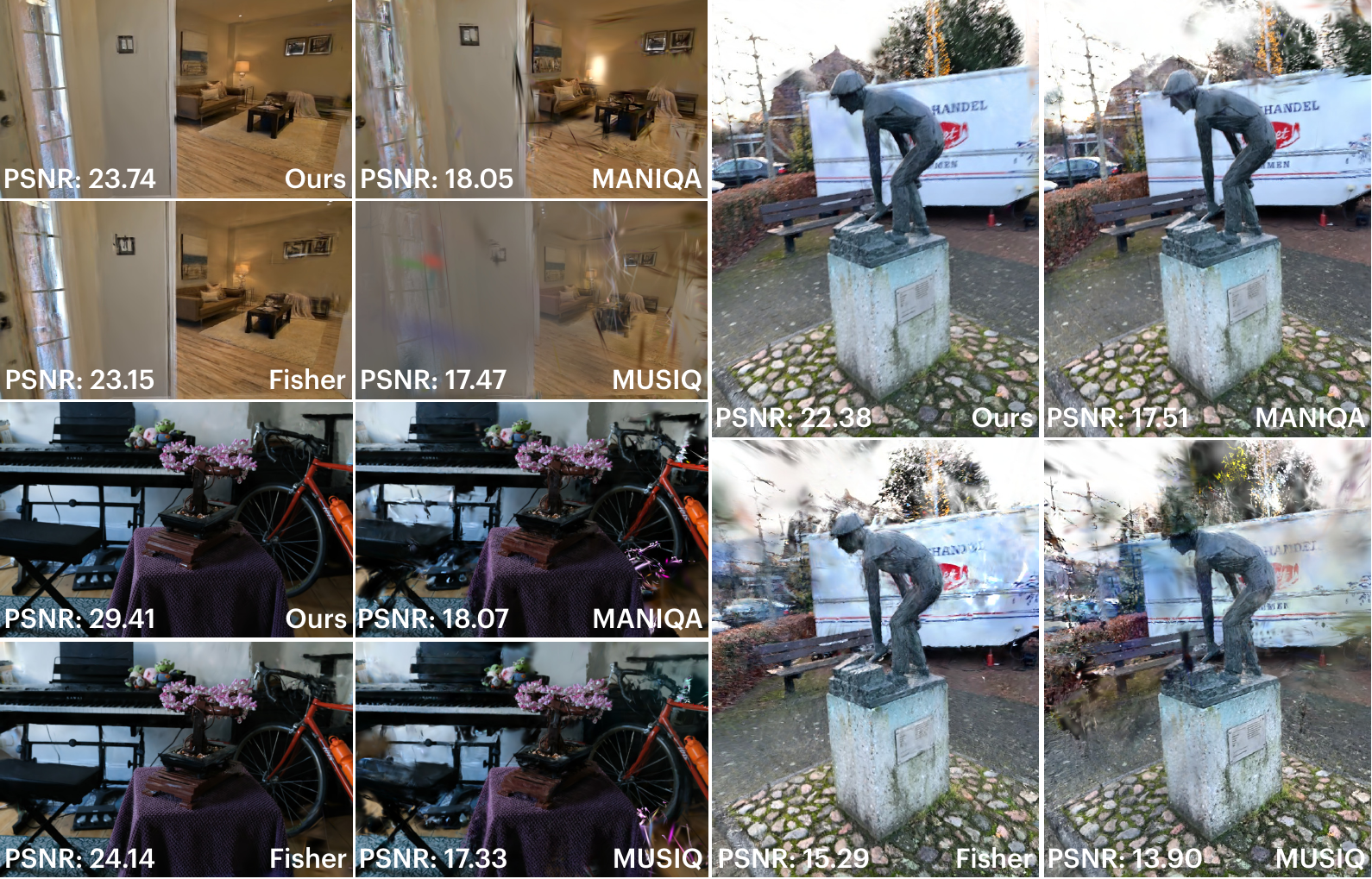}
    \caption{
        \textbf{Example NVS results after running active view selection with Gaussian Splatting on three datasets: RealEstate10K, Map-free Relocalisation, and Mip-NeRF360.}
        We compare our method with the state-of-the-art FisherRF model and present results from view selection pipelines using two strong no-reference IQA metrics: MANIQA and MUSIQ.
    }
    \label{fig:exp:nvs}
    \vspace{-0.1cm}
\end{figure*}

\begin{table}[t]
\centering
\caption{
    \textbf{
        NVS image quality under various active view selection strategy on Mip-NeRF360.
    }
    \colorbox[HTML]{F4CCCC}{\strut Quality}\colorbox[HTML]{FCE5CD}{\strut from}\colorbox[HTML]{D9EAD3}{\strut high}\colorbox[HTML]{C9DAF8}{\strut to}\colorbox[HTML]{D9D9D9}{\strut low}.
}
\label{tab:exp:nvs}
\resizebox{\columnwidth}{!}{%
\begin{tabular}{llccc}
\toprule
                            & Method      & PSNR $\uparrow$               & SSIM $\uparrow$              & LPIPS $\downarrow$           \\ \midrule
Oracle                      & FVS         & 20.92                         & 0.69                         & 0.34                         \\ \midrule
                            & ActiveNeRF  & 17.89                         & 0.53                         & 0.41                         \\
\multirow{-2}{*}{3D-based}  & FisherRF    & \cellcolor[HTML]{D9EAD3}20.34 & \cellcolor[HTML]{D9EAD3}0.60 & \cellcolor[HTML]{D9EAD3}0.37 \\ \midrule
                            & TOPIQ       & 19.52                         & \cellcolor[HTML]{D9D9D9}0.58 & \cellcolor[HTML]{C9DAF8}0.38 \\
                            & TRES        & 19.38                         & 0.57                         & \cellcolor[HTML]{C9DAF8}0.38 \\
                            & MANIQA      & \cellcolor[HTML]{D9D9D9}19.54 & \cellcolor[HTML]{C9DAF8}0.59 & \cellcolor[HTML]{D9EAD3}0.37 \\
                            & MUSIQ       & \cellcolor[HTML]{C9DAF8}19.62 & \cellcolor[HTML]{D9D9D9}0.58 & \cellcolor[HTML]{C9DAF8}0.38 \\
                            & NIMA        & 18.47                         & 0.56                         & \cellcolor[HTML]{D9D9D9}0.39 \\
                            & BRISQUE     & 18.64                         & 0.55                         & 0.40                         \\
                            & NIQE        & 18.10                         & 0.54                         & 0.41                         \\
                            & PIQE        & 18.03                         & 0.53                         & 0.41                         \\
                            & Random      & 17.91                         & 0.56                         & 0.43                         \\ \cmidrule{2-5} 
                            & Ours-DINOv2 & \cellcolor[HTML]{F4CCCC}21.11 & \cellcolor[HTML]{F4CCCC}0.65 & \cellcolor[HTML]{F4CCCC}0.32 \\
\multirow{-11}{*}{2D-based} & Ours-RepViT & \cellcolor[HTML]{FCE5CD}20.97 & \cellcolor[HTML]{FCE5CD}0.62 & \cellcolor[HTML]{FCE5CD}0.34 \\ \bottomrule
\end{tabular}%
}
\end{table}

\begin{table}[t]
\centering
\caption{\textbf{Active NVS results on RealEstate10K (RE10K) and Map-free Relocalisation (MFR) dataset.}}
\label{tab:supp:nvs_re10k_mfr}
\resizebox{\columnwidth}{!}{%
\begin{tabular}{lccclccc}
\toprule
         & \multicolumn{3}{c}{RE10K}                              &  & \multicolumn{3}{c}{MFR}                                \\ \cmidrule(lr){2-4} \cmidrule(l){6-8} 
         & PSNR $\uparrow$ & SSIM $\uparrow$ & LPIPS $\downarrow$ &  & PSNR $\uparrow$ & SSIM $\uparrow$ & LPIPS $\downarrow$ \\ \midrule
MANIQA   & 18.65           & 0.65            & 0.32               &  & 17.05           & 0.49            & 0.34               \\
MUSIQ    & 18.32           & 0.64            & 0.33               &  & 17.39           & 0.49            & 0.34               \\
FisherRF & 18.86           & 0.66            & 0.32               &  & 17.21           & 0.48            & 0.35               \\
Ours     & \textbf{19.29}  & \textbf{0.68}   & \textbf{0.29}      &  & \textbf{17.56}  & \textbf{0.50}   & \textbf{0.33}      \\ \bottomrule
\end{tabular}%
}
\end{table}
We set up the active novel view synthesis (NVS) task using the 3D Gaussian Splatting (3DGS) framework. Each system starts with four training views selected via farthest viewpoint sampling (FVS), then incrementally adds views until reaching 20 in total.

\boldstart{Scheduling.}  
We train the 3DGS model for $30$k iterations following its default configuration. Over these $30$k iterations, we add new views at specific intervals: $[400$, $900$, $1500$, $2200$, $3000$, $3900$, $4900$, $6000$, $7200$, $8500$, $9900$, $11400$, $13000$, $14700$, $16500$, $18400]$. This schedule, adapted from FisherRF, provides a final set of 20 training views, including the initial views.

\boldstart{3D Approaches.}  
For FisherRF, we use the official implementation. For the 3DGS adaptation of ActiveNeRF, we take the reported numbers from FisherRF.

\boldstart{2D Approaches.}  
All 2D image-based view selection methods follow the same pipeline, differing only in the IQA metric used. At each step, we render all unselected candidate views and apply the IQA metric to each; the lowest-quality image is chosen as the next training view. NR-IQA uses only the rendered image, while CrossScore treats it as a query and compares it to 5 reference views from the training views. All images are resized to a long side of 518.

\boldstart{Metrics.}  
For evaluating the novel view synthesis task, we use classical metrics, including PSNR, SSIM, and LPIPS.

\boldstart{Results.}
We conduct experiments on all three datasets, RE10K, MFR, and Mip-NeRF360, demonstrating that our active view selection method enables more realistic image rendering results compared to the state-of-the-art FisherRF, ActiveNeRF, and all no-reference IQA-based view selection methods. 
This result demonstrates the potential for a more streamlined and efficient active view selection pipeline.

These findings are supported by PSNR, SSIM, and LPIPS metrics, as shown in \cref{tab:exp:nvs,tab:supp:nvs_re10k_mfr}. Additionally, \cref{fig:exp:nvs} visualizes sample image renderings from our method, FisherRF, and active NVS systems using two strong NR-IQA baselines: MANIQA and MUSIQ. 

Note that our method not only outperforms all previous view selection approaches, but several 2D image-based view selection strategies also perform competitively with 3D-based methods. Compared to ActiveNeRF and FisherRF, which estimate uncertainty and information gain within 3D space, the 2D baselines achieve comparable image rendering quality without requiring modifications to 3D internal states, indicating a promising direction for analysing scene uncertainty from 2D rendered images.

\begin{figure*}[t]
    \centering
    \includegraphics[width=\textwidth]{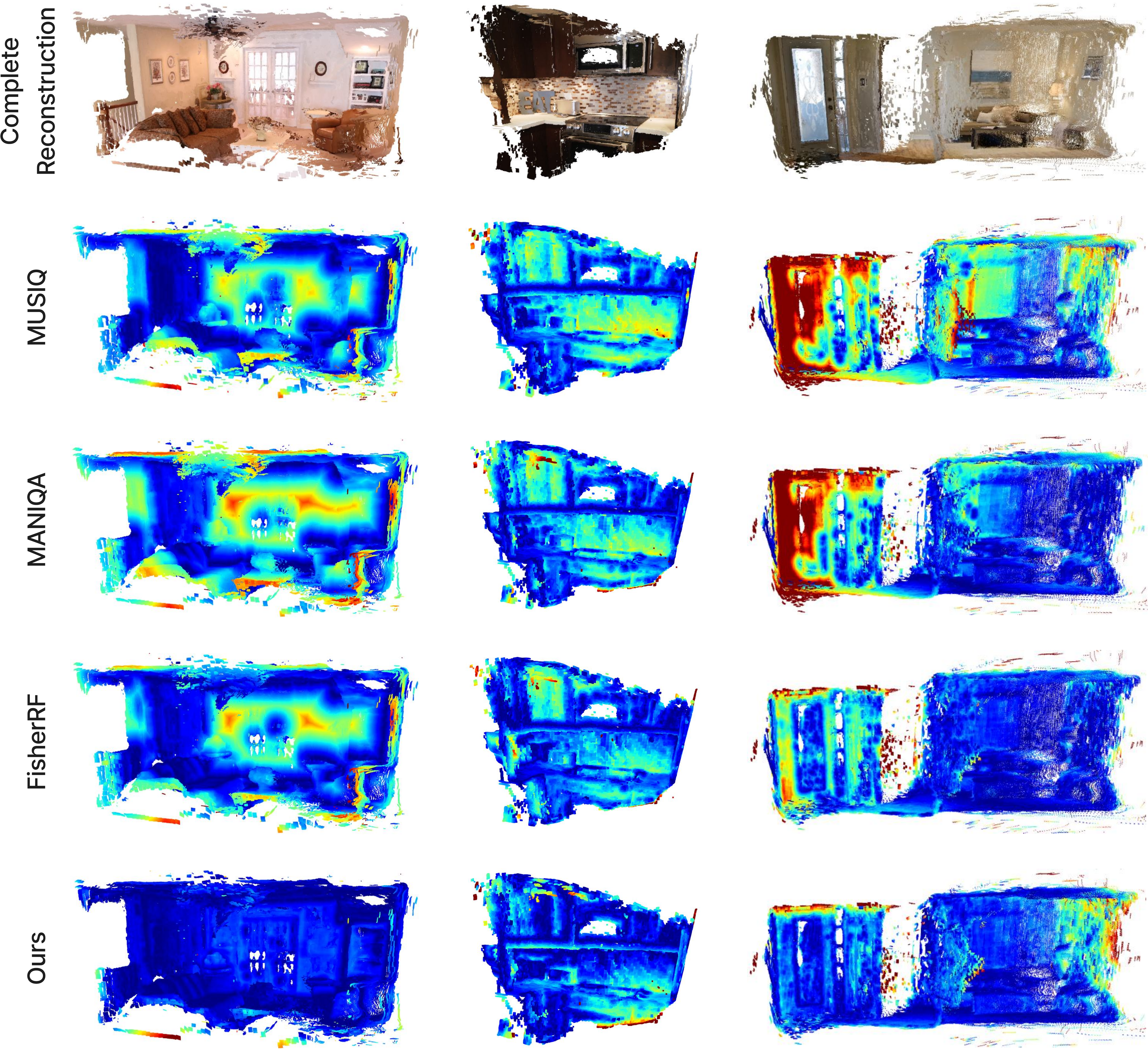}
    \caption{
        \textbf{Visualization of Scene Coverage.} We compare reconstruction errors for each view selection strategy. Errors are visualized as distances from each point in the “complete” point cloud to its nearest neighbor in the reconstruction produced by the subset of views selected by each method. \textcolor{blue}{\textbf{Blue}} areas indicate low reconstruction error, while \textcolor{red}{\textbf{red}} areas indicate higher errors due to limited view coverage.
    }
    \vspace{-0.2cm}
    \label{fig:exp:3d_recon}
\end{figure*}

\subsection{Task B: Scene Coverage}
\label{sec:exp:task_b_3d_recon}
Beyond NVS quality, we assess how well the selected views capture scene geometry in two settings: SfM with MASt3R~(\cref{sec:exp:task_b_3d_recon_mast3r}) and Active SLAM with SplaTAM~(\cref{sec:exp:task_b_3d_recon_active_slam}). MASt3R focuses on mapping accuracy, while Active SLAM prioritizes online performance.

\subsubsection{Scene Coverage in SfM}
\label{sec:exp:task_b_3d_recon_mast3r}
We use the pre-trained MASt3R~\cite{leroy2024grounding} model to reconstruct scenes from both the full set of views and selected subsets from each method. The full reconstruction serves as the reference. This evaluation reveals whether the selected views offer sufficient geometric coverage for high-quality reconstruction. Subsets are chosen in the NVS setup using various baselines by incrementally selecting 20 images from an initial set of 4 views during active NVS.

\boldstart{Metrics.}
We evaluate coverage using two complementary metrics. 
\textbf{Surface Coverage Ratio (SCR)}: The percentage of the reference model's surface area that is well-reconstructed in the subset model. This is computed as the percentage of reference points whose nearest distance is within a distance threshold, $\tau=0.01\times \text{reference scene extent}$.
\textbf{F-score}, following \cite{Knapitsch2017}, measures the bidirectional alignment between the complete point cloud and subset reconstructions, averaged across thresholds of 0.001, 0.01, and 0.1, providing a more comprehensive coverage summary than SCR.

\boldstart{Results.}
\Cref{tab:coverage-realestate} presents quantitative results on the RE10K dataset~\cite{zhou2018stereo}. 
Our cross-referencing method outperforms others in both SCR and F-score with a limited view budget of 20. This is further illustrated in \cref{fig:exp:3d_recon}, where we visualize reconstruction error heatmaps. Each heatmap shows the distance from points in the “complete” point cloud to their nearest neighbors in the reconstructed subset. In \cref{fig:exp:3d_recon}, our method (last row) has the fewest red zones, indicating stronger spatial coverage with fewer views. This efficiency benefits tasks like robot navigation and spatial analysis, where accurate 3D reconstruction is essential.

\begin{table}[t]
\centering
\caption{Scene coverage evaluation on the RE10K~\cite{zhou2018stereo} dataset in the SfM setup.}
\label{tab:coverage-realestate}
\begin{tabular}{lccc}
\toprule
Method     & SCR (\%) $\uparrow$   & F-score $\uparrow$  \\ 
\midrule
FisherRF   & 50.88 & 0.52 \\
MANIQA     & 51.87 & 0.52 \\
MUSIQ      & 44.08 & 0.47 \\
Ours       & \textbf{53.89} & \textbf{0.54} \\ 
\bottomrule
\end{tabular}
\end{table}

\subsubsection{Scene Coverage in Active-SLAM}
\label{sec:exp:task_b_3d_recon_active_slam}
We conduct an experiment in a 3DGS-based SLAM system using the Habitat simulator~\cite{habitat19iccv, puig2023habitat3, szot2021habitat2} to evaluate scene coverage. The robot explores Gibson~\cite{xiazamirhe2018gibsonenv} scenes with paths guided by different path/view selection functions. Building on SplaTAM~\cite{keetha2024splatam}, we use 3D Gaussians for scene representation and frontier-based exploration~\cite{yamauchi1997frontier} to propose paths. Viewpoints are sampled based on robot mobility and selected using 3D information gain or 2D image quality. Gaussians are converted to point clouds via their means before computing metrics. All steps follow the FisherRF evaluation protocol.

\boldstart{Metrics.}
Scene coverage and reconstruction quality are evaluated using two metrics.
First, Surface Coverage Ratio (SCR), as in the SfM setting but with a 5cm threshold per the FisherRF convention.
Second, Depth Mean Absolute Error (MAE) in meters, computed from depth images rendered at 1000 uniformly sampled poses.

\boldstart{Results.} 
As shown in \cref{tab:exp:active_slam}, our method achieves higher scene coverage ratio, lower depth error, and is significantly faster than the previous state of the art, FisherRF.

\begin{table}[t]
\centering
\caption{Scene coverage evaluation on Habitat-Gibson dataset with Active-SLAM setup.}
\label{tab:exp:active_slam}
\begin{tabular}{lccc}
\toprule
         & SCR (\%) $\uparrow$ & Depth MAE (m) $\downarrow$ & PSNR $\uparrow$ \\ \midrule
FisherRF & 92.89          & 0.092       & 22.58 \\
Ours     & \textbf{93.71}          & \textbf{0.076}       & \textbf{23.9}  \\ \bottomrule
\end{tabular}%
\end{table}

\subsection{Runtime Analysis}
\label{sec:exp:run_time}
\begin{figure}[t]
    \centering
    \includegraphics[width=1\columnwidth]{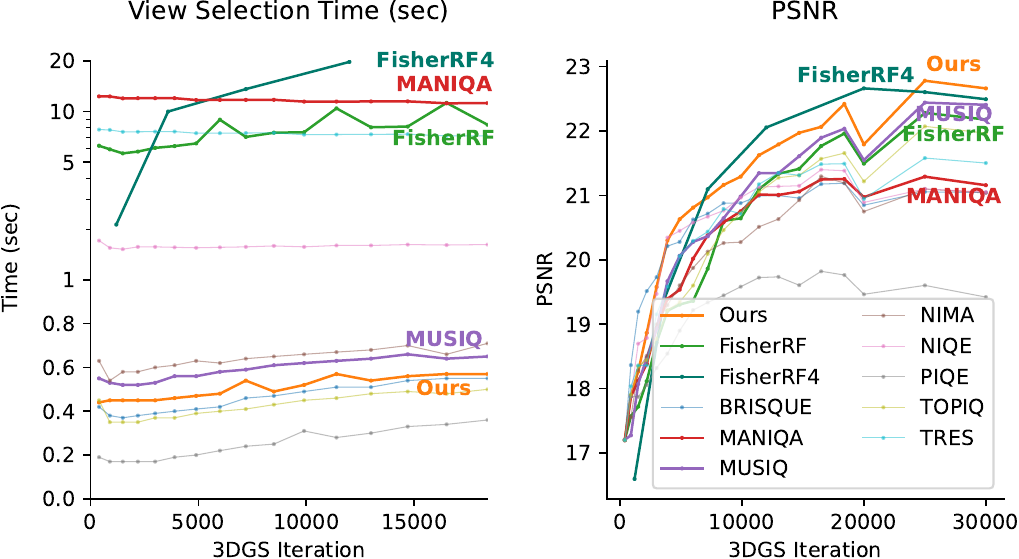}
    \caption{
        \textbf{
        View selection time (\emph{left}) and test split PSNR (\emph{right}) on the \emph{Garden} scene from the Mip-NeRF360 dataset.}
        Our method provides the highest NVS quality whilst being 14$\times$ faster than the state-of-the-art model FisherRF and 33$\times$ faster than the batched version FisherRF4.
        Strong baselines are highlighted with heavier line weights.
        Note that the upper part of the time axis, from 1 to 20 seconds, is plotted on a log scale. 
    }
    \label{fig:exp:time-psnr}
\end{figure}

We conducted a runtime analysis alongside an assessment of image quality rendering performance, with results presented in \cref{fig:exp:time-psnr,tab:exp:psnr-time-gpu}. Our method achieves the highest novel view synthesis (NVS) quality among all evaluated approaches, while maintaining low view selection time and small GPU memory footprint.
All experiments in this section were conducted using an Nvidia 4090 GPU.

\boldstart{Comparing with FisherRF.}
Our method is 14$\times$ faster than the state-of-the-art FisherRF and 33$\times$ faster than its batched 4-view selection variant, FisherRF4 while requiring less than half the GPU memory (\cref{fig:exp:time-psnr,tab:exp:psnr-time-gpu}). 
FisherRF~\cite{jiang2025fisherrf} reports a view \textit{evaluation} rate of 70 fps. Evaluating a typical Mip-NeRF360 scene requires assessing about 200 candidate views to select the next best view, leading to a per-view \textit{selection} time of 5–8 seconds.

\boldstart{Ours-DINOv2 vs Ours-RepViT.}
Of the two variants, Ours-DINOv2 yields slightly higher PSNR (+0.09) but suffers from high memory use and slower speed, limiting its practicality on low-budget platforms like robots.
Using the lightweight RepViT backbone~\cite{wang2024repvit}, we cut memory use by two-thirds and doubled the speed, making our method suitable for power-constrained GPUs.

\begin{table}[t]
\centering
\caption{
\textbf{PSNR, time, and peak GPU memory usage on \textit{Garden}.} Compared to the 3D-based uncertainty method FisherRF, the Cross-Reference IQA-based active view selection achieves higher PSNR with significantly lower frame selection time. Replacing DINOv2~\cite{oquab2023dinov2} backbone with the lightweight RepViT~\cite{wang2024repvit} backbone further reduces both selection time and memory usage by half, with only a minimal drop in performance.
}
\label{tab:exp:psnr-time-gpu}
\resizebox{\columnwidth}{!}{%
\begin{tabular}{lccc}
\toprule
Method    & PSNR $\uparrow$ & Time (s) $\downarrow$ & GPU Mem (GB) $\downarrow$ \\ \midrule
FisherRF    & 21.33           & 8.34                  & 15.8                      \\
Ours-DINOv2 & \textbf{23.05}  & 1.21                  & 20.1                      \\
Ours-RepViT      & 22.96           & \textbf{0.59}         & \textbf{8.3}              \\ \bottomrule
\end{tabular}%
}
\end{table}

\boldstart{Comparing with NR-IQA.}
Several strong no-reference IQA view selection strategies, including MANIQA~\cite{yang2022maniqa} and MUSIQ~\cite{ke2021musiq}, are also highlighted in \cref{fig:exp:time-psnr}. 
However, both methods require longer view selection times and underperform compared to our approach.

\boldstart{Non-constant View Selection Time.}
It is worth mentioning that view selection time is not constant but rather increases as the 3D Gaussian Splatting (3DGS) training progresses (\cref{fig:exp:time-psnr}). This is caused by the growing number of Gaussian primitives added during training, which leads to longer rendering time for IQA-based methods, and longer analysis time for 3D-based methods.

\section{Conclusion}
\label{sec:conclusion}
We reframe active view selection for 3D reconstruction as a 2D image quality assessment (IQA) task, improving NVS and 3D reconstruction performance while cutting costs by 14–33× over FisherRF.
Our cross-reference IQA model selects informative views without costly 3D uncertainty estimation and remains agnostic to 3D representations. This 2D quality-based approach enables faster, more efficient reconstruction for resource-limited systems like robotics and embedded applications.

\Boldstart{Acknowledgement.}
Zirui is supported by an ARIA Research Gift Grant from Meta Reality Lab, and Yash is supported by EPSRC AIMS CDT EP/S024050/1 and AWS.
{
    \small
    \bibliographystyle{ieeenat_fullname}
    \bibliography{main}
}
\maketitlesupplementary
\appendix

\section{Dataset Details}
We employ 3 datasets in our primary experiments. 
The first dataset, Map-Free Relocalisation (\textbf{MFR})~\cite{arnold2022mapfree}, originally developed for camera localization benchmarks, has been adapted for our network training. MFR contains 460 outdoor videos of various objects and buildings at a resolution of $540 \times 960$. The second dataset is \textbf{Mip-NeRF360}~\cite{barron2021mip}, which includes 9 videos capturing 360-degree scans of diverse scenes, both outdoor and indoor. Originally in approximately 4K resolution, all images are downscaled to $1066\times1600$ following similar practice in 3DGS. The third dataset is a selection of 10 videos randomly chosen from the RealEstate10K (\textbf{RE10K})~\cite{zhou2018stereo} dataset. These videos, originally at $1920 \times 1080$ resolution, are downscaled to $960 \times 540$ for easier 3DGS training.
We select these three datasets for their high photorealistic quality and widespread adoption in the field.
\textbf{Training and Testing Split:}
Our image scoring network is trained exclusively on the MFR dataset, from which we randomly select 348 videos for training and 11 for evaluation. In addition to evaluating on the MFR test split, we further assess the performance of our method on the Mip-NeRF360 and RE10K datasets.

\section{Detail Tables}
\subsection{Runtime and PSNR.}
We provide the exact time and PSNR measurement we presented at \cref{fig:teaser,fig:exp:time-psnr} in \cref{tab:supp:time_psnr_detail}.
Our method offers state of the art rendering quality after active view selection while maintaining low view selection time.

\begin{table}[t]
\centering
\caption{
    Runtime and PSNR detail on the \emph{garden} scene from Mip-NeRF360 dataset.
    Sort order:
    \colorbox[HTML]{F4CCCC}{\strut \quad}\colorbox[HTML]{FCE5CD}{\strut \quad}\colorbox[HTML]{D9EAD3}{\strut \quad}\colorbox[HTML]{C9DAF8}{\strut \quad}\colorbox[HTML]{D9D9D9}{\strut \quad}
}
\label{tab:supp:time_psnr_detail}
\resizebox{0.9\columnwidth}{!}{%
\begin{tabular}{llcc}
\toprule
                           & Method    & Time (sec)                   & PSNR                          \\ \midrule
Oracle                     & FVS       & 0.00                         & 23.15                         \\ \midrule
                           & FisherRF  & 8.34                         & 21.33                         \\
\multirow{-2}{*}{3D-based} & FisherRF4 & 19.70                        & \cellcolor[HTML]{D9EAD3}22.31 \\ \midrule
                           & TOPIQ     & \cellcolor[HTML]{FCE5CD}0.50 & 21.24                         \\
                           & TRES      & 7.17                         & \cellcolor[HTML]{D9D9D9}21.77 \\
                           & MANIQA    & 11.21                        & 21.61 \\
                           & MUSIQ     & \cellcolor[HTML]{D9D9D9}0.65 & \cellcolor[HTML]{C9DAF8}22.18 \\
                           & NIMA      & 0.71                         & 20.93                         \\
                           & BRISQUE   & \cellcolor[HTML]{D9EAD3}0.55 & 21.04                         \\
                           & NIQE      & 1.62                         & 21.24                         \\
                           & PIQE      & \cellcolor[HTML]{F4CCCC}0.36 & 19.45                         \\ \cmidrule(l){2-4} 
                           & Ours-DINOv2      & 1.21 & \cellcolor[HTML]{F4CCCC}23.05 \\
\multirow{-10}{*}{2D-based} & Ours-RepViT      & \cellcolor[HTML]{C9DAF8}0.59 & \cellcolor[HTML]{FCE5CD}22.96 \\
\bottomrule
\end{tabular}%
}
\end{table}

\subsection{Active NVS Quality on Mip-NeRF360.}
We provide per scene results for Active NVS experiment on Mip360 (\cref{tab:exp:nvs}) at \cref{tab:supp:nvs_mip360}

\begin{table*}[t]
\centering
\caption{Per scene Active NVS results on Mip-NeRF360 dataset, supporting \cref{tab:exp:nvs}.}
\label{tab:supp:nvs_mip360}
\resizebox{\textwidth}{!}{%
\begin{tabular}{@{}lcccccccccc@{}}
\toprule
\multicolumn{11}{c}{\cellcolor[HTML]{EFEFEF}\textbf{PSNR $\uparrow$}}                                                                                 \\ \midrule
                        & \multicolumn{10}{c}{Active View Selection Methods}                                                               \\ \cmidrule(l){2-11} 
\multirow{-2}{*}{Scene} & FisherRF      & Ours           & TOPIQ & TRES  & MANIQA        & MUSIQ & NIMA          & BRISQUE & NIQE  & PIQE  \\ \midrule
Bicycle                 & 18.44         & \textbf{18.56}          & 18.18 & 16.20 & 18.27         & 17.78 & 16.79         & 17.60   & 16.20 & 16.80 \\
Bonsai                  & 22.40         & \textbf{23.22} & 22.53 & 21.52 & 21.52         & 22.23 & 19.27         & 19.55   & 18.43 & 19.33 \\
Counter                 & 21.86         & \textbf{21.94} & 20.36 & 20.94 & 20.36         & 20.90 & 18.98         & 20.16   & 18.64 & 18.98 \\
Flowers                 & 15.46         & \textbf{16.16} & 15.04 & 15.36 & 15.31         & 15.44 & 14.89         & 14.76   & 14.00 & 14.70 \\
Garden                  & 21.33         & \textbf{22.96} & 21.24 & 21.77 & 21.61         & 22.18 & 20.93         & 21.04   & 21.24 & 19.45 \\
Kitchen                 & 23.60         & \textbf{23.89} & 22.63 & 22.88 & 23.25         & 23.54 & 20.06         & 19.33   & 20.60 & 19.18 \\
Room                    & 23.38         & \textbf{23.64} & 21.49 & 21.30 & 21.13         & 20.47 & 20.25         & 19.74   & 20.20 & 19.50 \\
Stump                   & 19.46         & \textbf{20.92} & 17.58 & 17.63 & 17.59         & 17.61 & 17.85         & 18.33   & 17.67 & 17.70 \\
Treehill                & 17.17         & \textbf{17.45} & 16.63 & 16.86 & 16.82         & 16.40 & 17.22         & 17.28   & 15.88 & 16.62 \\ \midrule
Mean                    & 20.34         & \textbf{20.97}          & 19.52 & 19.38 & 19.54         & 19.62 & 18.47         & 18.64   & 18.10 & 18.03 \\ \midrule
\multicolumn{11}{c}{\cellcolor[HTML]{EFEFEF}\textbf{SSIM $\uparrow$}}                                                                                 \\ \midrule
                        & \multicolumn{10}{c}{Active View Selection Methods}                                                               \\ \cmidrule(l){2-11} 
\multirow{-2}{*}{Scene} & FisherRF      & Ours           & TOPIQ & TRES  & MANIQA        & MUSIQ & NIMA          & BRISQUE & NIQE  & PIQE  \\ \midrule
Bicycle                 & \textbf{0.41} & 0.40           & 0.39  & 0.32  & 0.41          & 0.37  & 0.39          & 0.38    & 0.32  & 0.34  \\
Bonsai                  & 0.81          & \textbf{0.83}  & 0.82  & 0.79  & 0.78          & 0.80  & 0.71          & 0.70    & 0.69  & 0.70  \\
Counter                 & 0.76          & \textbf{0.77}  & 0.73  & 0.74  & 0.72          & 0.74  & 0.68          & 0.71    & 0.67  & 0.68  \\
Flowers                 & 0.31          & \textbf{0.35}  & 0.31  & 0.31  & 0.33          & 0.31  & 0.31          & 0.30    & 0.30  & 0.32  \\
Garden                  & 0.61          & \textbf{0.71}  & 0.63  & 0.64  & 0.64          & 0.65  & 0.63          & 0.63    & 0.66  & 0.54  \\
Kitchen                 & 0.80          & \textbf{0.83}  & 0.80  & 0.81  & 0.81          & 0.82  & 0.71          & 0.72    & 0.76  & 0.70  \\
Room                    & 0.80          & \textbf{0.82}  & 0.77  & 0.77  & 0.77          & 0.76  & 0.75          & 0.72    & 0.72  & 0.71  \\
Stump                   & 0.43          & \textbf{0.49}  & 0.38  & 0.38  & 0.38          & 0.38  & 0.39          & 0.41    & 0.38  & 0.38  \\
Treehill                & 0.42          & \textbf{0.43}  & 0.42  & 0.40  & 0.44          & 0.40  & 0.44          & 0.42    & 0.38  & 0.39  \\ \midrule
Mean                    & 0.60          & \textbf{0.62}  & 0.58  & 0.57  & 0.59          & 0.58  & 0.56          & 0.55    & 0.54  & 0.53  \\ \midrule
\multicolumn{11}{c}{\cellcolor[HTML]{EFEFEF}\textbf{LPIPS $\downarrow$}}                                                                                \\ \midrule
                        & \multicolumn{10}{c}{Active View Selection Methods}                                                               \\ \cmidrule(l){2-11} 
\multirow{-2}{*}{Scene} & FisherRF      & Ours           & TOPIQ & TRES  & MANIQA        & MUSIQ & NIMA          & BRISQUE & NIQE  & PIQE  \\ \midrule
Bicycle                 & 0.44          & 0.44           & 0.45  & 0.50  & \textbf{0.43} & 0.46  & 0.48          & 0.45    & 0.51  & 0.49  \\
Bonsai                  & 0.29          & \textbf{0.28}  & 0.29  & 0.31  & 0.32          & 0.30  & 0.35          & 0.36    & 0.38  & 0.36  \\
Counter                 & \textbf{0.30} & \textbf{0.30}  & 0.33  & 0.32  & 0.33          & 0.33  & 0.36          & 0.35    & 0.37  & 0.36  \\
Flowers                 & 0.50          & \textbf{0.48}  & 0.51  & 0.51  & 0.49          & 0.52  & 0.51          & 0.51    & 0.52  & 0.50  \\
Garden                  & 0.28          & \textbf{0.22}  & 0.27  & 0.26  & 0.26          & 0.25  & 0.28          & 0.28    & 0.26  & 0.34  \\
Kitchen                 & 0.24          & \textbf{0.20}  & 0.23  & 0.22  & 0.23          & 0.21  & 0.29          & 0.28    & 0.25  & 0.30  \\
Room                    & 0.31          & \textbf{0.29}  & 0.33  & 0.34  & 0.32          & 0.34  & 0.33          & 0.38    & 0.37  & 0.37  \\
Stump                   & 0.45          & \textbf{0.42}  & 0.49  & 0.48  & 0.49          & 0.49  & 0.48          & 0.47    & 0.49  & 0.49  \\
Treehill                & 0.47          & \textbf{0.45}  & 0.48  & 0.49  & 0.47          & 0.50  & \textbf{0.45} & 0.48    & 0.51  & 0.49  \\ \midrule
Mean                    & 0.37          & \textbf{0.34}  & 0.38  & 0.38  & 0.37          & 0.38  & 0.39          & 0.40    & 0.41  & 0.41  \\ \bottomrule
\end{tabular}%
}
\end{table*}

\section{Additional Active NVS Experiment on Out-of-Distribution Egocentric Data \& Future Directions}
One promising direction for future research is integrating our low-latency view selection method into applications requiring real-time performance, such as SLAM (Simultaneous Localization and Mapping) and AR/VR systems. For example, our method could guide a user or robot to capture images from target positions that achieve specific objectives, such as improved image quality in novel view synthesis (NVS) or enhanced accuracy and coverage in 3D reconstruction tasks. These scenarios demand not only high-quality view selection but also rapid decision-making to maintain seamless interactions and workflows.

State-of-the-art methods like FisherRF typically require 5–10 seconds for view selection, causing delays that force robots or users to pause movement while waiting for the system to compute the next optimal view. In contrast, our method significantly reduces this processing time to as little as 0.5 seconds for view selection. When factoring in the additional reaction time needed by users or robots, our approach enables seamless and non-blocking operation, offering substantial practical benefits for active vision tasks.

To explore the feasibility of this approach, we conducted an extra experiment integrating our view selection system into the active NVS task using the egocentric ARIA Digital Twin Catalogue (ARIA-DTC) dataset. Below, we detail the dataset, experimental setup, results, and potential future directions.

\begin{figure}[t]
    \centering
    \includegraphics[width=1\columnwidth]{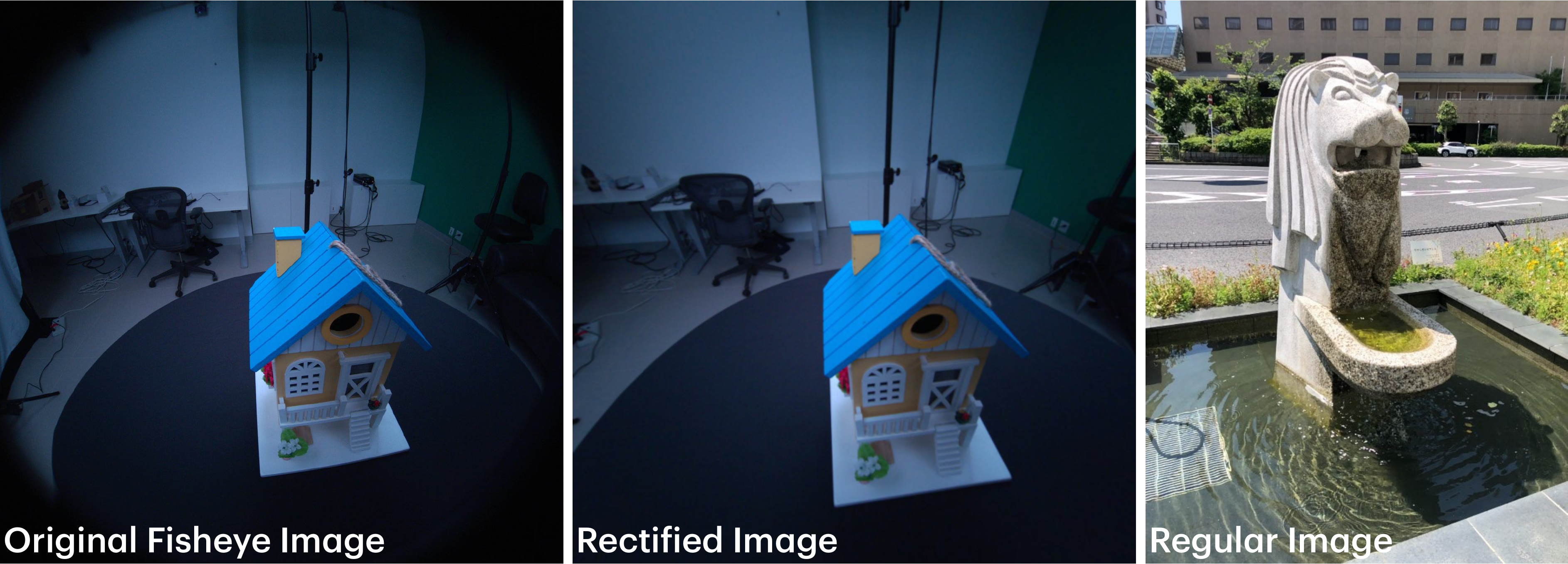}
    \caption{
        \textbf{Images Captured from Egocentric Devices (ARIA) vs. Mobile Phones.}
        Note that even after rectification, egocentric images differ in three key ways: 1) vignette effect, 2) lower exposure, and 3) a wider field of view. 
    }
    \label{fig:supp:aria}
\end{figure}

\Boldstart{Dataset.}
We utilized the \href{https://www.projectaria.com/datasets/dtc/}{ARIA Digital Twin Catalog} (ARIA-DTC) dataset, an egocentric dataset comprising video sequences captured using smart glasses. The video is captured by a human user walking around a table while observing an object placed on it, while wearing the glasses. 

This dataset poses unique challenges for view selection due to the distinct camera characteristics of smart glasses, which differ significantly from typical mobile phone cameras. Key factors include the super-wide field of view (FoV), measuring $110^\circ$ before rectification and $95^\circ$ after, compared to $71^\circ$ for a typical 26mm mobile phone lens. Additionally, the fisheye lens causes a persistent vignette effect, and the dataset features a unique camera color profile and exposure control unlike standard imaging devices.
\Cref{fig:supp:aria} provides an example image from the dataset, highlighting these distinctive characteristics. The original image resolution is $1408 \times 1408$ pixels, which we rectified and resized to $707\times707$ pixels for 3DGS training.

\Boldstart{Experimental Setup.}
From the ARIA-DTC dataset, we randomly selected 10 videos as a representative subset for our experiments. These videos are used to evaluate the generalisation capability of our method on the out-of-distribution egocentric data in the active NVS task. 
We employed the same 3D Gaussian Splatting-based active NVS system described in \cref{sec:exp:setup}, adhering to the same scheduling and procedures detailed in the main paper. The only difference in this setup was the use of the ARIA-DTC dataset.

\Boldstart{Results.}
The results of the active NVS task on the ARIA-DTC dataset, summarized in \cref{tab:supp:nvs_aria}, demonstrate that our method achieves comparable performance to the state-of-the-art FisherRF model while being over 10 times faster. Our method delivers perceptually similar results, as indicated by identical LPIPS scores, with only slight reductions in PSNR (by 0.22) and SSIM (by 0.01). FisherRF achieves this slight quality advantage through its use of 3D Gaussian primitives to evaluate uncertainty, enhancing robustness to variations in camera models and characteristics. However, this comes at the significant cost of computational speed, limiting its usage in real-time applications.
\begin{table}[t]
\centering
\caption{
    \textbf{Active NVS results on ARIA-DTC dataset.}
    Despite the dramatic image characteristics difference between ARIA devices and mobile phones (\cref{fig:supp:aria}), our view selection method generalizes well to egocentric settings, matching FisherRF even though our method is trained only on mobile phone images.
    With its rapid view selection speed and potential for further enhancement through fine-tuning, our method shows great promise for integration into real-time active vision applications such as SLAM and AR/VR.
}
\label{tab:supp:nvs_aria}
\begin{tabular}{lccc}
\toprule
         & PSNR $\uparrow$           & SSIM $\uparrow$          & LPIPS $\downarrow$         \\ \midrule
FisherRF & \textbf{21.07} & \textbf{0.80} & \textbf{0.34} \\
Ours     & 20.85          & 0.79          & \textbf{0.34} \\ \bottomrule
\end{tabular}%
\end{table}

\Boldstart{Future Directions.}
While not being trained on the egocentric data, our method generalizes well to the ARIA-DTC dataset. 
This suggests two future directions.
\textit{First}, fine-tuning on egocentric datasets could further enhance its performance and adaptability.
\textit{Second}, integration into real-time applications. 
The low latency and high accuracy aspects of our view selection method make it a practical choice for real-time systems, such as SLAM and AR/VR, where rapid decision-making is essential. 
Overall, these findings highlight the balance our approach achieves between speed and quality, making it especially useful for challenging egocentric environments in real world.

\end{document}